\documentclass[letterpaper]{article} 
\usepackage{aaai2026}  
\usepackage{times}  
\usepackage{helvet}  
\usepackage{courier}  
\usepackage[hyphens]{url}  
\usepackage{graphicx} 
\usepackage{natbib}  
\usepackage{caption} 
\usepackage{amsfonts}
\usepackage{booktabs}
\usepackage{amssymb}
\usepackage{colortbl}
\usepackage{multirow}
\usepackage{xcolor}
\usepackage{amsmath}
\usepackage{algorithm}
\usepackage{algorithmicx}
\usepackage{algpseudocode}
\newcommand{\meanstd}[2]{#1 ({\footnotesize$\pm$ #2})}
\usepackage{newfloat}
\usepackage{listings}
\DeclareCaptionStyle{ruled}{labelfont=normalfont,labelsep=colon,strut=off} 
\floatstyle{ruled}
\newfloat{listing}{tb}{lst}{}
\floatname{listing}{Listing}
\usepackage{hyperref}

\title{ReLayout: Integrating Relation Reasoning for Content-aware Layout Generation with Multi-modal Large Language Models}
\author{
    Jiaxu Tian\textsuperscript{\rm 1}\equalcontrib,
    Xuehui Yu\textsuperscript{\rm 2}\equalcontrib,
    Yaoxing Wang\textsuperscript{\rm 1}\equalcontrib,
    Pan Wang\textsuperscript{\rm 2},
    Guangqian Guo\textsuperscript{\rm 1},
    Shan Gao\textsuperscript{\rm 1}\thanks{Corresponding author.}
}
\affiliations{
    \textsuperscript{\rm 1}Northwestern Polytechnical University, \textsuperscript{\rm 2}Tencent \\
    jiaxutian@neau.edu.cn, \{xuehuiyu, reluwang\}@tencent.com,\\
    \{wangyx24, guogq21\}@mail.nwpu.edu.cn,
 gaoshan@nwpu.edu.cn
}

\usepackage{bibentry}

\begin{document}

\maketitle
\begin{abstract}
Content-aware layout aims to arrange design elements appropriately on a given canvas to convey information effectively. 
Recently, the trend for this task has been to leverage large language models (LLMs) to generate layouts automatically, achieving remarkable performance.
However, existing LLM-based methods fail to adequately interpret spatial relationships among visual themes and design elements, leading to structural and diversity problems in layout generation.
To address this issue, we introduce ReLayout, a novel method that leverages relation-CoT to generate more reasonable and aesthetically coherent layouts by fundamentally originating from design concepts.
Specifically, we enhance layout annotations by introducing explicit relation definitions, such as region, saliency, and margin between elements, with the goal of decomposing the layout into smaller, structured, and recursive layouts, thereby enabling the generation of more structured layouts. 
Furthermore, based on these defined relationships, we introduce a layout prototype rebalance sampler, which defines layout prototype features across three dimensions and quantifies distinct layout styles. This sampler addresses uniformity issues in generation that arise from data bias in the prototype distribution balance process.
Extensive experimental results verify that ReLayout outperforms baselines and can generate structural and diverse layouts that are more aligned with human aesthetics and more explainable.
\end{abstract}

\begin{links}
    \link{Datasets}{https://huggingface.co/datasets/jiaxutian/ReLayout}
\end{links}

\section{Introduction}
\label{sec:intro}
Layout is an essential part of graphic design, aiming to convey information through the appropriate arrangement of elements such as logos and texts. Due to its importance, layout has various applications, spanning scenarios like documents~\cite{document1,document2}, UIs~\cite{UI1,UI2}, magazines~\cite{magazines1,magazines2} and posters~\cite{poster1,poster2}. Among these, when the main visual element flows into an application, such as advertising posters, achieving harmony between the arrangement of elements and the canvas becomes one of the key goals. We call layout generation under the above condition content-aware layout generation. 

This field is particularly challenging because it requires the integration of design elements, such as logos and text, with visual content to produce layouts that are both usable and aesthetically pleasing. Furthermore, the model needs to generate diverse layouts to ensure diversity.
To address these challenges, researchers have proposed various methods~\cite{contentgan,ralf,pku,cgl} based on generative models~\cite{gan, vae, diffusion} to enhance the quality of generated layouts. Among these methods, RALF~\cite{ralf}, as a transformer-based~\cite{transformer} method, has achieved notable advancements. It adopts a retrieval augmentation method to mitigate the data scarcity problem. Nevertheless, it treats layout generation only as a numerical problem, failing to capture the semantics, which prevents the model from generating visually and textually coherent layouts. 

\begin{figure}[t]
	\includegraphics[width=\columnwidth]{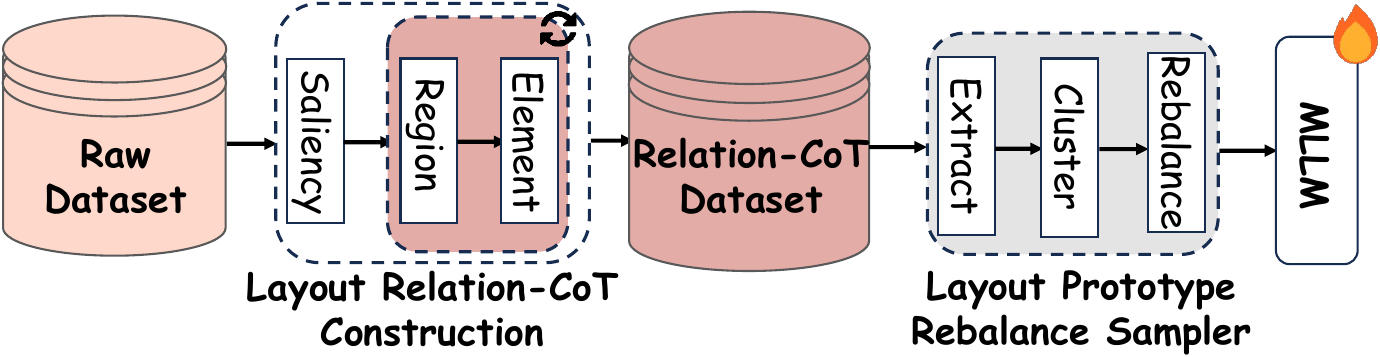}
	\caption{Pipeline of ReLayout. We adopt the layout relation-CoT construction to add relation annotations on raw datasets. Then we use the layout prototype rebalance sampler to adjust the distribution of the new dataset for training.} \label{fig:pipeline}
\end{figure}

Recently, two LLM-based methods~\cite{layoutprompter,posterllama} have emerged, aiming to leverage the ability of large language models to generate high-quality layouts. 
For instance, LayoutPrompter~\cite{layoutprompter} employs dynamic exemplar selection to generate layouts without requiring training but cannot take a canvas image as input, thereby missing out on a significant amount of information.
PosterLlama~\cite{posterllama}, as the current SOTA, trains a MLLM to generate visually and textually coherent layouts. 
However, these methods remain limited to outputting coordinate information at the element-level (e.g., "where to place" individual elements) and focusing only on layout-level outcomes, lacking the structural-level organization of \textbf{element relations} that bridges element-level positioning with layout-level design concepts. This limitation leads to two critical issues in layout generation: 
(1) structural problem, where related elements fail to maintain proper spatial relationships, as illustrated in Figure~\ref{fig:error}(a), where PosterLlama produces overlapping elements, incorrect alignments, and fails to capture parallel relationships; and (2) diversity problem, where the generated layouts lack the rich structural variation found, as shown in Figure~\ref{fig:error}(b), where these methods, without explicit modeling of element relationships, degrade to similar structural arrangements.
To address these issues, we propose ReLayout, a content-aware layout generation framework based on a MLLM, drawing inspiration from how designers organize layouts through structural element relations. Our core contribution lies in explicitly modeling design logic through a CoT reasoning mechanism~\cite{cot} that deciphers element relations. As illustrated in the layout relation-CoT construction in Figure~\ref{fig:pipeline}, it decomposes layouts into recursive, nested hierarchical structures (e.g., tree representations) by defining a relation space encompassing salient, region, and element. This structured approach enhances the model’s ability to generate semantically coherent layouts by leveraging relations between elements. Additionally, we introduce the layout prototype rebalance sampler, which quantifies the layout prototype into a three-dimensional feature space of saliency, region, and margin between elements based on the layout relation-CoT construction. By integrating feature clustering with weighted sampling, the sampler mitigates the long-tail distribution problem in the dataset, enabling balanced learning of diverse layout prototypes.
User studies and visualization demonstrate that ReLayout outperforms state-of-the-art methods, achieving significant improvements in usability and diversity.
In summary, our contributions are as follows:
\begin{itemize} 
    \item We propose ReLayout, a relation-CoT paradigm designed to address hierarchical layout design challenges, specifically tackling structural and diversity problems via explicit spatial relations and layout prototype balancing.
    \item We introduce a layout relation-CoT construction mechanism that decomposes layout element relationships into a hierarchical structure while incorporating element relation annotations into existing layout datasets.
    \item We develop a layout prototype rebalance sampler, which quantifies layout prototypes through feature clustering and employs weighted sampling to ensure adaptability across diverse real-world scenarios.
    \item We propose two datasets enriched with more layout information based on the layout relation-CoT construction, which we will release publicly to the community.
\end{itemize}

\section{Related Work}
\subsection{Automatic Layout Generation}
\textbf{Content-agnostic:} Content-agnostic layout generation aims to create layouts independent of specific content. LayoutGAN~\cite{li2019layoutgan} is the first method to introduce GAN for addressing this task; in addition, approaches involving VAE~\cite{jiang2022coarse,jyothi2019layoutvae} or Diffusion models~\cite{chai2023layoutdm,zhang2023layoutdiffusion,inoue2023layoutdm} have also been employed to solve content-agnostic layout generation tasks. LayoutNUWA~\cite{layoutnuwa} is an LLM-based method that has achieved good performance using HTML format. This also demonstrates that LLMs have advantages over other generative methods in layout generation tasks.

\textbf{Content-aware:} Content-aware layout generation not only focuses on the quality of the generated layout like content-agnostic layout generation but also considers the harmony between the layout and the canvas. ContentGAN~\cite{contentgan} is the first to tackle the above problem. Starting from CGL-GAN~\cite{cgl}, subsequent works mostly begin leveraging saliency maps. DS-GAN~\cite{pku} uses a CNN-LSTM model to balance graphic and content-aware metrics. RADM~\cite{radm} is the first diffusion-based method to incorporate textual content into layout tasks. RALF~\cite{ralf} leverages a retrieval augmentation method to mitigate the data scarcity problem. Thanks to the power of LLMs, LayoutPrompter~\cite{layoutprompter} and PosterLlama~\cite{posterllama} demonstrate remarkable capabilities in the field of layout generation. The former achieves a training-free approach by selecting prompt examples with constraint layouts similar to test samples. The latter, PosterLlama, trains an adapter and fine-tunes the model to generate coherent visual and textual layouts. Among these works, LLM-based methods have become the mainstream method, with PosterLlama, the current SOTA, demonstrating outstanding performance. 

However, they fail to capture the rich relationships between elements. In contrast, our method explicitly represents these relationships and decomposes the layout into smaller, structured, and recursive layouts. This leads to a layout that is both more visually appealing and more explainable.
\subsection{Multi-modal Large Language Models}
\textbf{Advancements:} LLMs have demonstrated remarkable capabilities in natural language understanding with billions of parameters. Based on this, MLLMs have achieved remarkable progress by integrating cross-modal data including visual, auditory, and other sensory data streams~\cite{blip,clip}, thereby significantly expanding their range of applications, such as GPT-4~\cite{gpt}, Gemini~\cite{gemini}, and Claude 3, as well as open-source models like InternVL~\cite{internvl2.5} and LLaVA-OneVision~\cite{llavaov}. These models have been widely applied across diverse fields, including healthcare~\cite{healthcare,healthcare1} and agriculture~\cite{agriculture,agriculture1}.

\textbf{Techniques:} In recent years, several techniques have enhanced LLM capabilities. Few-shot learning~\cite{fewshot} allows models to adapt to new tasks with minimal examples, reducing the need for large datasets. Chain-of-thought (CoT) prompting~\cite{cot} improves reasoning by guiding models to break down complex problems step by step. LoRA fine-tuning~\cite{lora} efficiently adapts models by adding small trainable matrices to specific layers, reducing memory and computation costs while maintaining strong performance.

In our work, we leverage InternVL as the base model and apply LoRA fine-tuning like PosterLlama to efficiently adapt it to layout generation tasks. Moreover, motivated by CoT, we improve our output format to guide the model in generating more reasonable and explainable layouts. Experimental results demonstrate the effectiveness of ReLayout.

\section{Methods}
\subsection{Overview}


Given a set of constraints, our goal is to generate a well-arranged layout. A layout $\mathcal{L}$ can be represented as a set of $N$ elements: $\mathcal{L}=\{\mathbf{e}_1,\dots,\mathbf{e}_N\}=\{(c_1, \mathbf{b}_1),\dots,(c_N, \mathbf{b}_N)\}$, where each element $e_i$ consists of its class $c_i$ and corresponding bounding box $\mathbf{b}_i=[x_i, y_i, w_i, h_i]$. In our work, multi-modal inputs are a canvas image $\mathbf{C}$ and foreground elements $\mathcal{F}=\{(\mathbf{t}_i, \mathbf{p}_i)\}^{N}_{i=1}$, where $\mathbf{t}_i$ represents text (which can be empty) and $\mathbf{p}_i$ represents an element image (which can be empty except under condition constraints).

Furthermore, the constraints consist of two types: (1) content-aware constraints (avoiding occlusion of salient objects) and (2) user-specified constraints (e.g., generating bounding boxes conditioned on element categories).

\begin{figure*}[t]
	\includegraphics[width=\textwidth]{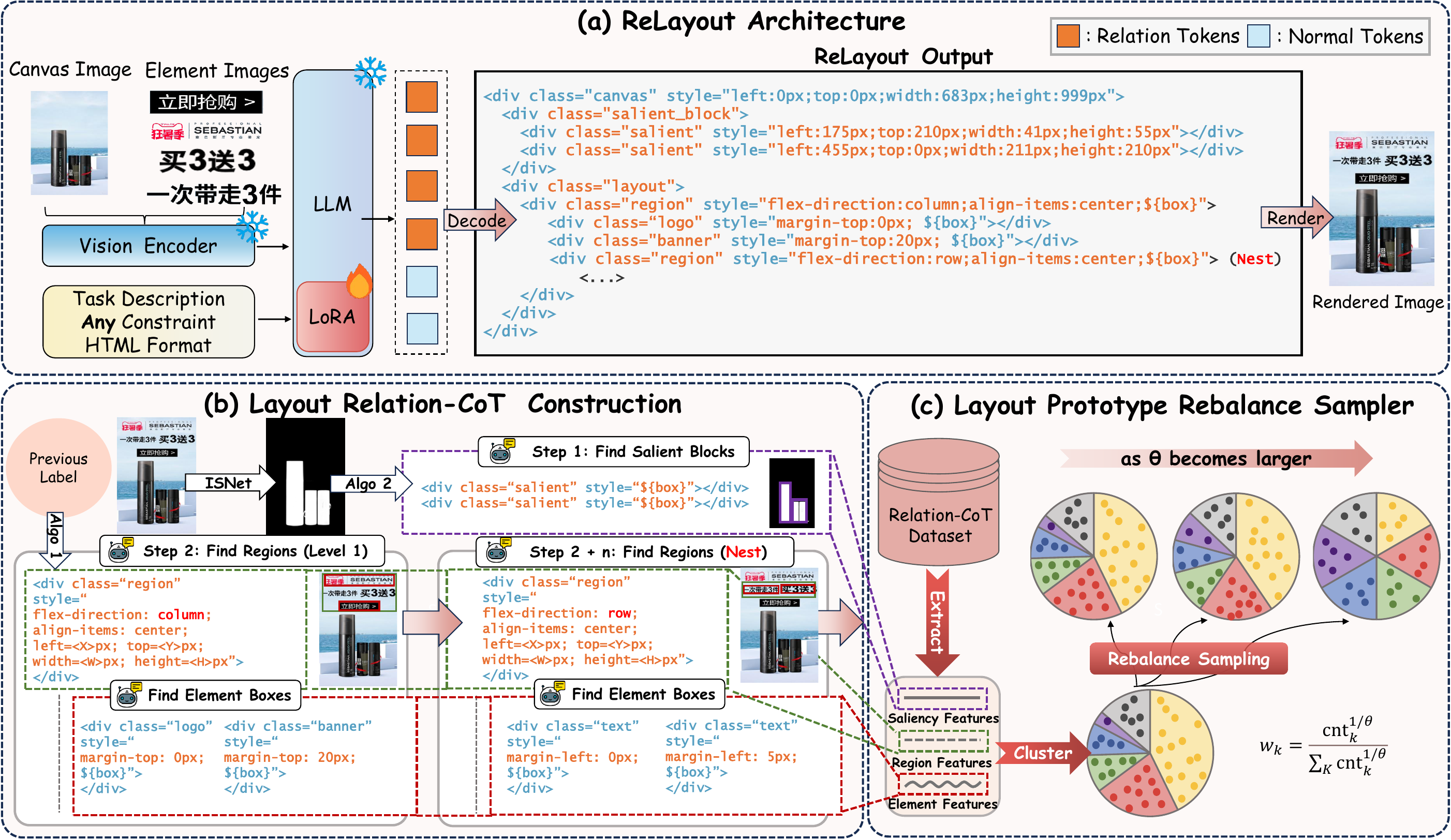}
	\caption{(a) is ReLayout training process and its output distinction from previous methods. The bottom part is two key components of ReLayout. (b) illustrates the relation labels construction logic. (c) represents the layout dataset resampling process, which adjusts the dataset distribution to achieve a more balanced layout dataset.} \label{fig:model}
\end{figure*}

The pipeline of ReLayout, shown in Figure~\ref{fig:pipeline}, consists of two key components: layout relation-CoT construction and layout prototype rebalance sampler. The layout relation-CoT construction explicitly models the layout relations from three aspects: margin between elements, region, and saliency. These relations will be used for the training of the MLLM to enhance the model's usability. Furthermore, these explicit relation models enable us to balance the samples in different clusters from the perspective of design styles, so as to achieve better optimization and diverse results.
The inference procedure is illustrated in Figure~\ref{fig:model}(a). Unlike previous layout generation methods based on LLMs to directly generate layout coordinates, our method first predicts the structured relations (highlighted in orange) and then generates the layout coordinates based on the provided canvas image $\mathbf{C}$ and foreground elements $\mathcal{F}$.

\subsection{Layout Relation-CoT Construction}


\renewcommand{\algorithmicrequire}{\textbf{Input:}}
\renewcommand{\algorithmicprocedure}{\textbf{Procedure:}}
\renewcommand{\algorithmicensure}{\textbf{Output:}}
To fully leverage the extensive knowledge of LLMs in layout design, we choose HTML to represent layouts. However, unlike previous LLM-based methods that represent layouts using HTML~\cite{layoutprompter,posterllama}, we introduce two types of relation spaces: \textbf{region} and \textbf{saliency} (see Figure~\ref{fig:model}(b)). These relational spaces are designed to address the shortcomings of previous methods, which often generate layouts that are poorly structured and lack human aesthetic appeal.
\begin{figure}[tb]
	\includegraphics[width=\columnwidth]{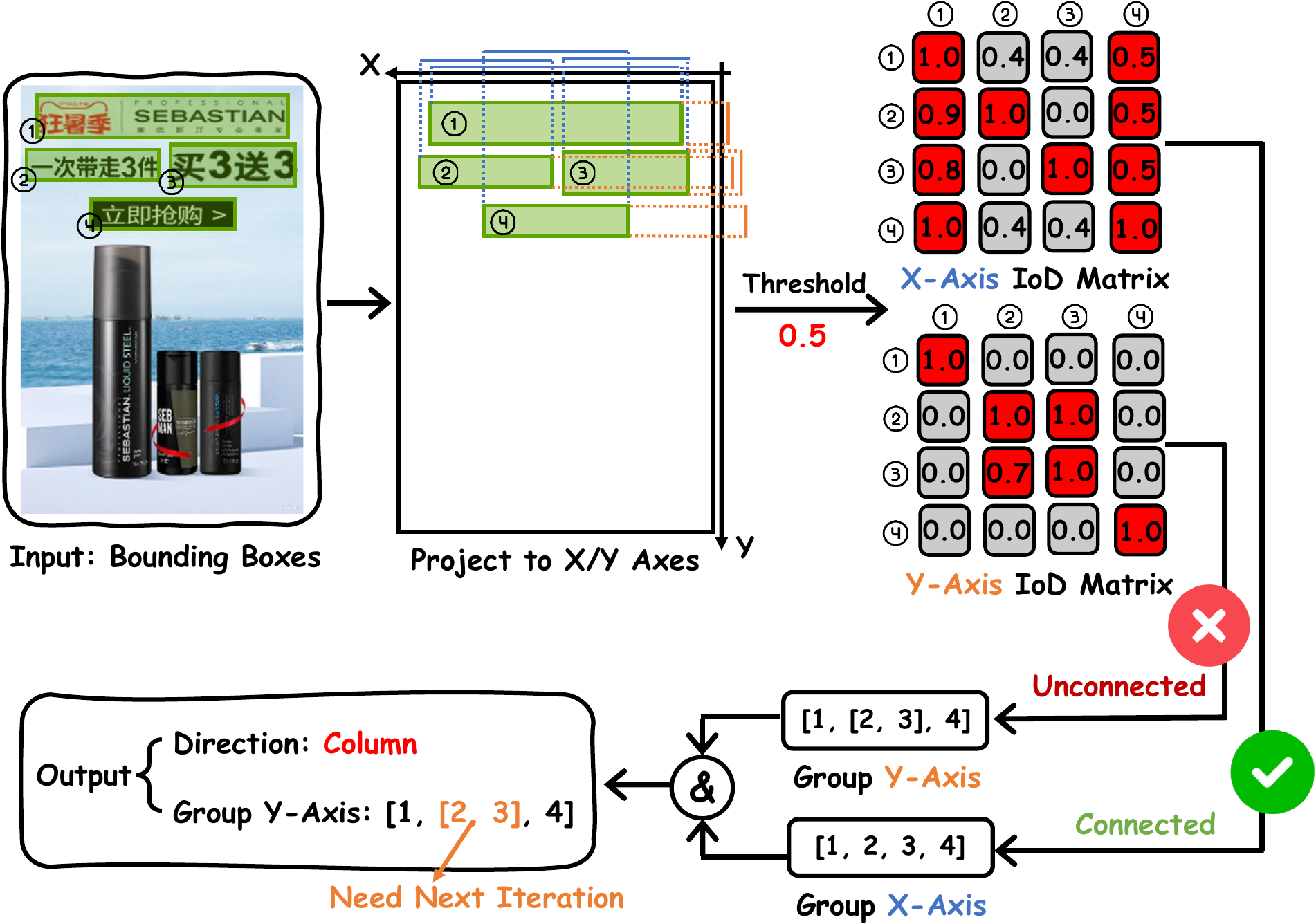}
	\caption{Examples of hierarchical decomposition of complex layouts based on different directions.} \label{fig:iter}
\end{figure}
\newline

\noindent\textbf{Region:} Caused by the fact that LLMs are inherently more sensitive to highly structured data, we introduce region. Region $\mathcal{R}$ serves as the fundamental unit of spatial arrangement, with its internal structure adhering to a single direction pattern. It can be understood as individual small layouts, similar to the structure of a tree. Thus it is both \textbf{nestable} and \textbf{recursive}. This makes the layout annotations formed by it highly structured, allowing the generation of complex overall arrangements through simple construction rules.

Region is defined by three key properties: $\mathcal{R}=\left(d, a, \mathbf{b}\right)$, where $d$ is the flex-direction, representing the arrangement direction of elements within the region: $d\in\left\{\textit{row},\textit{column}\right\}$, $a$ represents align-items, 
and $\mathbf{b}$ represents the region's position and size. As illustrated in Step 2 and Step $2+n$ of Figure~\ref{fig:model}(b), regions are constructed step by step. We use Algorithm~\ref{algo:complex} and Figure~\ref{fig:model}(b) as examples to describe the specific steps of constructing our region. (1) We first perform the x-axis and y-axis projection operations on each element of this level-1 region. (2) Using GroupByOverlap, we analyze the IoD (Intersection over Detection)~\cite{yu2020scale} matrix of projections to group bounding boxes into $G_x$ (x-axis groups) and $G_y$ (y-axis groups), where IoD is defined as the intersection between the detection box and the ignored region divided by the area of the detection box. (3) Based on the group counts and variances, we determine the layout direction. At this point, we have obtained the direction of the level-1 region in Step 2 of Figure~\ref{fig:model}(b). Finally, we only need to recursively apply this process to each group to further subdivide the region, constructing a hierarchical structure like Step 2 + n of Figure~\ref{fig:model}(b). Figure~\ref{fig:iter} illustrates the process of converting bounding boxes into nested structures enriched with layout information under the heuristic Algorithm~\ref{algo:complex}.


Furthermore, parallel $\mathcal{P}$ (see the second column of Figure~\ref{fig:error}(a)) is a specialized type of region, sharing the same fundamental attributes. It is typically employed for the parallel presentation of two or more related elements. These elements maintain uniform visual sizes and align along a designated axis (either row or column) to ensure consistency and symmetry within the layout.

For each element within a region, we introduce an additional attribute, \textit{margin}, to represent relative position, i.e., the spacing between elements. When the region is arranged in a row, this attribute is defined as \textit{margin-left}, whereas in a column, it is specified as \textit{margin-top}. Using this property, we can effectively control the overall layout compactness.

\begin{algorithm}[!ht]
\caption{Estimate Layout Direction}
\begin{algorithmic}[1]
\Require Bounding boxes $\mathcal{B}$, overlap threshold $\phi$.
\Ensure (Direction, $G$)
\State $L_x \gets$ project bounding boxes $\mathcal{B}$ to x-axis;
\State $L_y \gets$ project bounding boxes $\mathcal{B}$ to y-axis;
\State $G_x \gets$ GroupByOverlap($L_x$, $\phi$);
\State $G_y \gets$ GroupByOverlap($L_y$, $\phi$);
\If{$|G_x| = 1$ AND $|G_y| > 1$}
    \State\Return {(``column'', $G_y$)}
\ElsIf{$|G_y| = 1$ AND $|G_x| > 1$}
    \State\Return{(``row'', $G_x$)}
\Else
    \State $V_x \gets$ ComputeGroupVariance($G_x$)
    \State $V_y \gets$ ComputeGroupVariance($G_y$)
    \If{$V_x \leq V_y$}
        \Return {(``row'', $G_x$)}
    \Else 
        \State\Return {(``column'', $G_y$)}
    \EndIf
\EndIf
\Function{GroupByOverlap}{$L$, $\phi$}
    \State $edges \gets \emptyset$
    \For{each pair $(i,j)$ in $L$}
    \If{IoD($L[i]$, $L[j]$) $\geq\phi$ }
        \State $edges \gets edges \cup \{(i,j)\}$
    \EndIf
\EndFor
    \State $groups \gets$ FindConnectedComponents($edges$)
\State\Return $groups$
\EndFunction
\end{algorithmic}
\label{algo:complex}
\end{algorithm}

\noindent\textbf{Saliency:} Inspired by the goal that designers usually avoid placing elements over salient objects, we introduce salient blocks $\mathcal{S}$ to help the model better grasp their features intuitively. These blocks are represented as a series of bounding boxes and are seamlessly integrated into an HTML-based representation. To detect these salient blocks, we propose an iterative algorithm that efficiently identifies prominent areas through integral image computation. This algorithm, detailed in the supplementary materials, progressively selects non-overlapping rectangular regions by evaluating their saliency scores based on the density of white and black pixels, ensuring the captured regions align with natural visual attention patterns. This unified way allows the model to understand the spatial relationships between elements and the background more effectively. Moreover, Section~\ref{sec:ablation} also explains that adding salient blocks is crucial for the model to understand the background.

\noindent\textbf{Sequence formalization:} 
Our input sequence comprises a primary instruction, a task description (e.g., \texttt{"layout generation with given class"}), and an input HTML format. Four mask tokens (\texttt{<X>}, \texttt{<Y>}, \texttt{<W>}, \texttt{<H>}) are introduced to facilitate their prediction.

We combine Saliency and Region components to form a unified HTML format as the output sequence (refer to the ReLayout output shown in Figure~\ref{fig:model}(a)). This format provides an effective strategy for constructing relational CoT in LLM-based layout methods. Additionally, the CoT-annotated dataset generated on the PKU and CGL datasets will be open to the community for further research.
\subsection{Layout Prototype Rebalance Sampler}
Building upon layout relation-CoT annotations, we propose the layout prototype rebalance sampler to address the issue of limited diversity in previous methods. By the process, our method ensures a more even distribution across diverse layout prototypes, providing the model with greater opportunities to learn and generalize over a broader range of layouts. As shown in Figure~\ref{fig:model}(c), our layout prototype rebalance sampler consists of three key operations: feature extraction, feature clustering, and rebalance sampling. Below, we provide a detailed explanation of each operation.
\newline
\noindent\textbf{Feature extraction:} The \( i^{\text{th}} \) layout prototype is to be primarily characterized by three dimensions: $\left\{ \mathcal{S}_i, \mathcal{R}_i, \mathcal{E}_i \right\}$.

The set of saliency bounding boxes in the \( i^{\text{th}} \) layout is denoted as \( \mathcal{S}_i \), given by: $\mathcal{S}_i = \left\{ \mathbf{b}_{i,j}^\text{s} \right\}_{j=1}^{r_i}$. The number of saliency bounding boxes in layout $L_i$ is given by ${r_i}\in  \left\{1, 2, 3, 4\right\}$. The saliency feature vector for layout 
$L_i$ captures the weighted center of all saliency boxes. Specifically, the centroid coordinates are computed as the weighted average of geometric centers of the saliency boxes, where the weights are proportional to the area of each saliency box.

We define the set of regions in a layout as $\mathcal{R}_i = \{\mathbf{b}_{i,j}^\text{r}, d_{i,j}\}_{j=1}^{s_i}$, where $d_{i,j} \in \{\text{row}, \text{column}\}$ represents the region's alignment direction. Then, we extract statistical features from $\mathcal{R}_i$ to describe their spatial distribution.
It includes the total number of regions \( s_i \), the standard deviations of their centroid coordinates \( \sigma_{i}^\text{x} \) and \( \sigma_{i}^\text{y} \), and the counts of row-aligned and column-aligned regions, \( n_i^{\text{row}} \) and \( n_i^{\text{column}} \), to roughly quantify the overall layout structure.

We define the element set of the \( i^{\text{th}} \) layout as $\mathcal{E}_i = \{ c_{i,j}\}_{j=1}^{t_i}$, where $ t_i $ is the total number of elements, and \( c_{i,j} \) represents the category of the \( j^{\text{th}} \) element in the \( i^{\text{th}} \) layout. We believe that the layout is highly related to the types and numbers of elements. Therefore, we define element-level features as follows: $\mathbf{f}_i^\text{e} = \begin{pmatrix}\sum_{j=1}^{t_i} \mathbb{I}(c_{i,j} = c_k)
\end{pmatrix}_{k=1}^K$, 
where $\mathbf{f}_i^\text{e}$ encodes the frequency of each element category $c_k$ within the layout. Here, $K$ denotes the predefined number of element categories (e.g., text, logo) in the dataset.
\newline
\noindent\textbf{Feature cluster:} 
\label{sec:temperature}
The final feature representation is constructed by weighted concatenation of the three feature dimensions:
\begin{equation}
\mathbf{f}_i = \alpha \mathbf{f}_i^\text{s} \oplus \beta \mathbf{f}_i^\text{r} \oplus \gamma \mathbf{f}_i^\text{e}
\end{equation}

Using these aggregated feature vectors, we apply K-means clustering to group layouts with similar characteristics. We set the number of clusters $K=8$ to maintain reasonable group sizes for subsequent analysis. 


\noindent\textbf{Rebalance sampling:} After obtaining $K$ clusters, we introduce a weighted sampling strategy to balance each cluster's influence and prevent large clusters from dominating the training. Specifically, we assign a sampling weight to each cluster based on its size:
\begin{equation}
\mathbf{w} = \frac{\mathbf{cnt}^{1/\theta}}{\| \mathbf{cnt}^{1/\theta} \|_1},
\end{equation}
where $\| \mathbf{cnt}^{1/\theta} \|_1 = \sum_{k=1}^K \text{cnt}_k^{1/\theta}$ and $\text{cnt}_k$ represents the number of layouts in cluster $k$, and $\theta$ is a hyperparameter that controls the distribution of weights. Larger $\theta$ makes the weights more uniform, ensuring small clusters are sampled more. However, overly large $\theta$ may over-sample rare clusters, distorting the data distribution. Smaller $\theta$ gives higher weights to large clusters, preserving the original distribution. But this may under-sample small clusters, limiting the model’s ability to learn from rare cases.

\section{Experiments}


\label{sec:exp}
\subsection{Datasets}
We use two publicly available e-commerce datasets, CGL~\cite{cgl} and PKU~\cite{pku}. The PKU dataset includes three element categories: Logo, Banner, and Text, while the CGL dataset has an additional element category called Embellishment. 
CGL contains 60,548 annotated poster-layout pairs and 1,000 unannotated canvases. PKU consists of 9,974 annotated poster-layout pairs and 905 unannotated canvases. 
Notably, considering that when designing text (especially text that needs an underlay), designers often treat the text and its underlay as a single unified element. To better reflect the practical value of the work, the "Banner" refers to elements where Intersection over Union (IoU) or IoD~\cite{yu2020scale} between the text and its underlay is greater than 0.95. We evaluate all baselines based on the above setting of categories. 
Finally, due to PKU and CGL datasets not providing annotated poster validation and test splits, we approximately divide the datasets into train/validation/test sets with a ratio of 8:1:1. 
Additionally, we create an extra hard split for each dataset. This hard split is selected from the test and validation sets based on the following conditions: (1) one region is nested within another, (2) a parallel relationship, and (3) the number of elements exceeding four.

\begin{table*}[tb!]
\caption{Performance comparison of the C $\xrightarrow{}$ S + P layout generation task on the PKU and CGL datasets. The best result is highlighted in bold, the \underline{second-best result} is underlined, and the row corresponding to \colorbox{red!8}{our method} is marked in red.}
\Large
    \centering
    \resizebox{\textwidth}{!}{
    \begin{tabular}{lcccccccccc}
    \toprule
    \multirow{3}{*}{Method} & \multicolumn{5}{c}{Test Split}&\multicolumn{5}{c}{Hard Split}\\
    \cmidrule(lr){2-6}\cmidrule(lr){7-11}
    &\multicolumn{3}{c}{Graphic}& \multicolumn{2}{c}{Content}&\multicolumn{3}{c}{Graphic}& \multicolumn{2}{c}{Content} \\
    \cmidrule(lr){2-4} \cmidrule(lr){5-6} \cmidrule(lr){7-9} \cmidrule(lr){10-11}
     &$\Delta$Val$\downarrow$  & Ove$\downarrow$ &  FD$\downarrow$ & Rea$\downarrow$ & Occ$\downarrow$  &$\Delta$Val$\downarrow$  & Ove$\downarrow$ &  FD$\downarrow$ & Rea$\downarrow$ & Occ$\downarrow$ \\
\midrule
\multicolumn{11}{l}{\textbf{PKU Annotated Dataset}}\\
\textcolor{gray}{Real Data}
&\textcolor{gray}{\meanstd{0.0000}{0.0000}} & \textcolor{gray}{\meanstd{0.0035}{0.0000}} & \textcolor{gray}{-} & \textcolor{gray}{\meanstd{0.1545}{0.0000}} & \textcolor{gray}{\meanstd{0.0639}{0.0000}}&\textcolor{gray}{\meanstd{0.0000}{0.0000}} & \textcolor{gray}{\meanstd{0.0047}{0.0000}} & \textcolor{gray}{-} & \textcolor{gray}{\meanstd{0.1673}{0.0000}} & \textcolor{gray}{\meanstd{0.0387}{0.0000}}\\
LayoutPrompter
&\meanstd{0.0015}{0.0000} & \underline{\meanstd{0.0090}{0.0000}} &\meanstd{8.0392}{0.0000} &\meanstd{0.1683}{0.0000} &\meanstd{0.1452}{0.0000}&\meanstd{0.0632}{0.0000} & \underline{\meanstd{0.0170}{0.0000}} &\meanstd{16.7438}{0.0000} &\meanstd{0.1883}{0.0000} &\meanstd{0.1530}{0.0000}\\
RALF
& \textbf{\meanstd{0.0000}{0.0000}} & \meanstd{0.0915}{0.0023}& \meanstd{15.5497}{0.1499} & \meanstd{0.1617}{0.0005} & \meanstd{0.0866}{0.0024}
&\textbf{\meanstd{0.0000}{0.0000}} & \meanstd{0.1740}{0.0031} &\meanstd{26.7978}{0.2282} &\meanstd{0.1728}{0.0003} &\underline{\meanstd{0.0639}{0.0010}}\\
PosterLlama
&\meanstd{0.0002}{0.0003} & \meanstd{0.0211}{0.0018} & \meanstd{3.5318}{0.2160} & \meanstd{0.1612}{0.0002} & \underline{\meanstd{0.0863}{0.0019}}&\meanstd{0.0007}{0.0002} & \meanstd{0.0318}{0.0021} & \meanstd{5.9256}{0.1448} & \underline{\meanstd{0.1727}{0.0003}} & \meanstd{0.0659}{0.0006}\\
InternVL2.5-8B
& \meanstd{0.0054}{0.0009} & \meanstd{0.0175}{0.0004}& \underline{\meanstd{2.6175}{0.0905}} & \textbf{\meanstd{0.1588}{0.0003}} & \meanstd{0.0885}{0.0016}& \meanstd{0.0050}{0.0014} & \meanstd{0.0323}{0.0014}& \underline{\meanstd{4.3106}{0.2717}} & \textbf{\meanstd{0.1717}{0.0002}} & \meanstd{0.0661}{0.0025}\\
\rowcolor{red!8}ReLayout (Ours) 
& \underline{\meanstd{0.0001}{0.0002}}& \textbf{\meanstd{0.0086}{0.0011}}& \textbf{\meanstd{1.7865}{0.1195}}& \underline{{\meanstd{0.1600}{0.0004}}}& \textbf{\meanstd{0.0857}{0.0010}}& \underline{\meanstd{0.0004}{0.0005}}& \textbf{\meanstd{0.0109}{0.0001}}& \textbf{\meanstd{3.4615}{0.1304}}& \underline{\meanstd{0.1727}{0.0005}}& \textbf{\meanstd{0.0637}{0.0002}}\\
\bottomrule
\multicolumn{11}{l}{\textbf{CGL Annotated Dataset}}\\
\textcolor{gray}{Real Data}
& \textcolor{gray}{\meanstd{0.0000}{0.0000}} & \textcolor{gray}{\meanstd{0.0060}{0.0000}} & \textcolor{gray}{-} & \textcolor{gray}{\meanstd{0.1654}{0.0000}} & \textcolor{gray}{\meanstd{0.0771}{0.0000}}
&\textcolor{gray}{\meanstd{0.0000}{0.0000}} & \textcolor{gray}{\meanstd{0.0100}{0.0000}} & \textcolor{gray}{-} & \textcolor{gray}{\meanstd{0.1758}{0.0000}} & \textcolor{gray}{\meanstd{0.0540}{0.0000}}\\
LayoutPrompter
&\meanstd{0.0125}{0.0000} & \underline{\meanstd{0.0094}{0.0000}} &\meanstd{6.7951}{0.0000} &\meanstd{0.1787}{0.0000} &\meanstd{0.1510}{0.0000}
&\meanstd{0.0184}{0.0000} & \underline{\meanstd{0.0124}{0.0000}} &\meanstd{9.3699}{0.0000} &\meanstd{0.1932}{0.0000} &\meanstd{0.1313}{0.0000}\\
RALF
& \meanstd{0.0147}{0.0001} & \meanstd{0.0283}{0.0007} &\textbf{\meanstd{0.9277}{0.0312}} & \underline{\meanstd{0.1649}{0.0002}} & \textbf{\meanstd{0.0744}{0.0001}}
&\meanstd{0.0213}{0.0001} & \meanstd{0.0478}{0.0010} & \textbf{\meanstd{1.7152}{0.0557}} &\textbf{\meanstd{0.1760}{0.0002}} &\textbf{\meanstd{0.0518}{0.0001}}\\
PosterLlama
& \underline{\meanstd{0.0012}{0.0003}}  & \meanstd{0.0102}{0.0010} & \meanstd{4.4151}{0.0129} & \meanstd{0.1674}{0.0001} & \meanstd{0.0931}{0.0004} & \textbf{\meanstd{0.0017}{0.0004}} & \meanstd{0.0183}{0.0013}  & \meanstd{7.1272}{0.0236} & \meanstd{0.1799}{0.0003} & \meanstd{0.0747}{0.0005} \\
InternVL-2.5-8B 
& \meanstd{0.0062}{0.0005} & \meanstd{0.0114}{0.0007}& \meanstd{2.8395}{0.1031} & \underline{\meanstd{0.1649}{0.0002}} & \meanstd{0.0796}{0.0003}& \meanstd{0.0098}{0.0010} & \meanstd{0.0195}{0.0009}& \meanstd{4.3051}{0.0629} & \underline{\meanstd{0.1765}{0.0002}} & \meanstd{0.0588}{0.0007} \\
\rowcolor{red!8}ReLayout (Ours) 
& \textbf{\meanstd{0.0004}{0.0002}} & \textbf{\meanstd{0.0088}{0.0003}} &  \underline{\meanstd{1.9311}{0.0120}} & \textbf{\meanstd{0.1648}{0.0001}}  & \underline{\meanstd{0.0787}{0.0001}} & \underline{\meanstd{0.0023}{0.0001}} & \textbf{\meanstd{0.0117}{0.0006}} & \underline{\meanstd{3.1917}{0.0215}} & \textbf{\meanstd{0.1760}{0.0001}} & \underline{\meanstd{0.0580}{0.0004}} \\
\bottomrule
    \end{tabular}
    }
    \label{tab:big table}
\end{table*}

\subsection{Baselines}
We use the following three SOTA methods to compare our method. (1) LayoutPrompter~\cite{layoutprompter} employs a dynamic exemplar selection module to eliminate the need for LLM training. In our work, we use the GPT-3.5 turbo instruct model because the GPT-3 text-davinci-003 model mentioned in the original paper is unavailable. (2) RALF~\cite{ralf} uses a retrieval augmentation to address the data
scarcity issue. Unlike the original work, which filtered out posters with more than 10 elements for PKU, we extend the maximum number of elements to 20, enabling more complex layouts and ensuring a fairer comparison. (3) PosterLlama~\cite{posterllama} builds upon the architecture of a multi-modal large language model and trains an adapter to improve the accuracy of content-aware text layout generation.

\subsection{Implementation Details}
Our model is fine-tuned on InternVL2.5-8B~\cite{internvl2.5}, which utilizes the InternViT-300M~\cite{internvit} vision encoder and the InternLM2.5-7B~\cite{internlm} language model. Each experiment is conducted on eight NVIDIA A800 GPUs. We follow the settings specified in InternVL for training and inference by default.

\begin{figure}[tb]
	\includegraphics[width=\columnwidth]{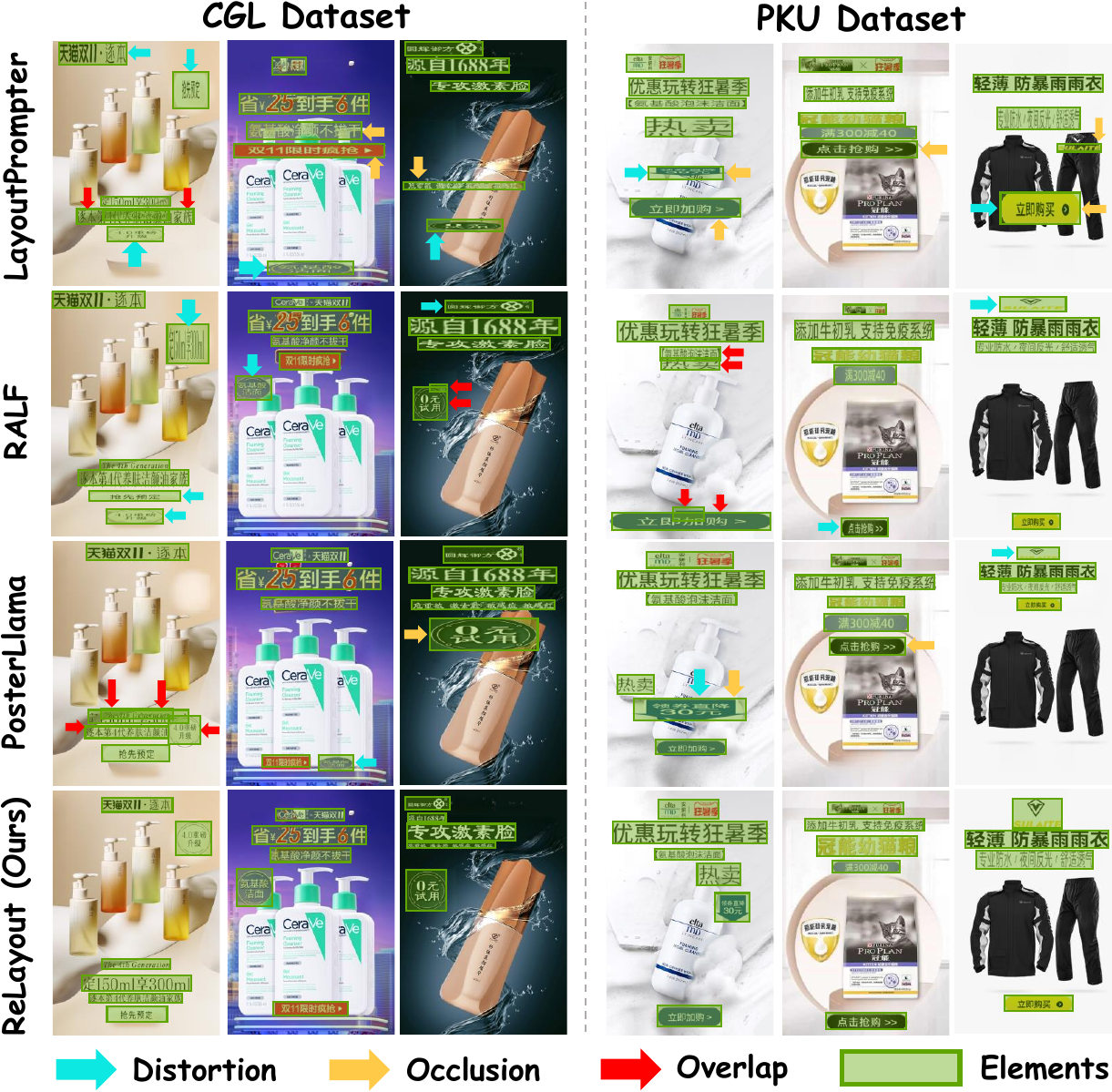}
	\caption{Qualitative comparison on the PKU and CGL datasets. Baselines layouts show noticeable errors, while ours meet basic requirements and better align with human aesthetics in margin and arrangement.} \label{fig:big fig}
\end{figure}

\subsection{Evaluation Metrics}
Following the evaluation metrics from previous works~\cite{cgl,fid,pku}, we apply five metrics. 
Additionally, we refine the overlap metric to ensure a more reasonable evaluation.
\newline
\noindent\textbf{Graphic metrics:} 
These metrics evaluate the graphic quality of the layout without considering the canvas. Validity (Val) represents the ratio of elements that are greater than 0.1\% of the canvas. All other metrics are calculated using only these valid elements. Due to the presence of small elements like embellishments in CGL, we use $\Delta$Val as a metric for evaluation. In previous works~\cite{ralf,posterllama,pku}, Overlap (Ove) is the average IOU across all element pairs. However, it has a notable limitation: when a layout contains a pair of completely overlapping elements along with many pairs that don’t overlap at all, the metric will fail to reflect the actual layout quality. On the other hand, if a canvas has a pair of elements with extensive overlap in the real world, it is considered a failure. Therefore, we use the maximum IoU to evaluate the generated layouts. We calculate Fréchet Distance (FD) in the feature space derived from bounding boxes and categories to evaluate overall layout quality.
\newline
\noindent\textbf{Content metrics:} 
These metrics assess harmony between the generated layout and the canvas. Occlusion (Occ) calculates the pixel coverage ratio of layout elements over saliency maps. Readability score (Rea) evaluates text clarity using average pixel gradients, where lower scores indicate clearer text.
\newline
\noindent\textbf{User study:} 
In the layout generation field, the current metrics are insufficient to fully evaluate the quality of a layout. Therefore, we conduct user studies. 

In terms of structure, we randomly select 300 images from the PKU dataset and invite 6 professional designers. For each image, we generate layouts using five different methods and present all layouts simultaneously in a shuffled order, with model names not perceived by users. Users assess each row of results based on two criteria: (1) identify all layouts that meet basic usability standards (e.g., no overlap, no occlusion), denoted as $P_\text{use}$; and (2) select the single best layout according to professional design principles, considering appropriate margin, relative size, distance from products and overall visual harmony, denoted as $P_\text{best}$.
\begin{table}[!ht]
\caption{User study on structural evaluation.}
\Large
    \centering
    \resizebox{\columnwidth}{!}{
    \begin{tabular}{lcccccc}
    \toprule
&LayoutPrompter&RALF & PosterLlama  & InternVL & \cellcolor{red!8}ReLayout\\
\midrule
$P_\text{use}$&36.0\%&50.3\%&71.3\%&78.3\%&\cellcolor{red!8}\textbf{91.0\%} \\
$P_\text{best}$&1.3\%&9.7\% &12.0\%&10.7\%&\cellcolor{red!8}\textbf{66.3\%} \\
\bottomrule
    \end{tabular}
}
    \label{tab:user study struct}
\end{table}
\begin{table}[!ht]
\caption{User study on diversity evaluation.}
\Large
    \centering
    \resizebox{\columnwidth}{!}{
    \begin{tabular}{lccccc}
    \toprule
&RALF & PosterLlama  & InternVL & \cellcolor{red!8}ReLayout\\
\midrule
$\text{Score}$&41&47&36&\cellcolor{red!8}\textbf{56} \\
$\mathbf{cnt}$&(18, 23, 9)&(14, 25, 11)&(21, 22, 7)&\cellcolor{red!8}\textbf{(11, 22, 17)} \\
\bottomrule
    \end{tabular}
}
    \label{tab:user study diverse}
\end{table}

In terms of diversity, we randomly select 50 images from the PKU dataset and invite 6 professional designers. LayoutPrompter is excluded due to poor usability, leaving four methods for evaluation, each run with three different random seeds (0, 1, 2). For each image, results from all methods are displayed simultaneously in a shuffled order, with model names not perceived by users to ensure unbiased assessment. Users are instructed to evaluate diversity based on differences in relative position (e.g., alignment) and text size—any variation in either aspect is considered a distinct style. Each row presents four methods (three images per method), and diversity is scored as 0 (one style), 1 (two styles), or 2 (three or more styles).

\subsection{Main Results}
Since designers usually design elements first before arranging the overall layout, our experiments primarily focus on generating the positions and sizes of elements based on given categories and the auxiliary information that each model can support as input.
\newline
\noindent\textbf{Quantitative comparison:} 
Table~\ref{tab:big table} presents a comparison of different methods on the test and hard split of the PKU and CGL datasets. It can be observed that the metrics of our method are either the best or the second-best. Specifically, on the PKU dataset, our method demonstrates the best performance on most metrics, with a particularly notable improvement in the Ove metric. On the CGL dataset, while our method does not achieve the best performance across all metrics, it consistently outperforms others on the Ove metric. Furthermore, when transitioning from the test split to the hard split, the degradation in our method's metrics is significantly smaller compared to other methods, highlighting the robustness of our approach. 
These improvements are attributed to the annotations margin property of our relation-CoT and the resampling strategy that effectively balances the dataset. These enhancements demonstrate that our method is better at generating more structured layouts. 
Although RALF performs well on the Occ and FD metrics in the CGL dataset, their higher Ove score reduces their practical usability, which is also reflected in the subsequent visualizations~\ref{fig:big fig}. 
Plus, Table~\ref{tab:user study struct} shows that ReLayout performs significantly better in aligning with human aesthetic preferences compared to baselines. Furthermore, Table~\ref{tab:user study diverse} demonstrates that our method also achieves the highest diversity, exhibiting a greater number and variety of distinct layout styles under different seed settings.

\begin{table*}[tb!]
\caption{Cross-dataset evaluation on PKU and CGL datasets.}
\Large
    \centering
    \resizebox{\textwidth}{!}{
    \begin{tabular}{lccccccc}
    \toprule
    {Train} & {Test} & {Method} &$\Delta$Val$\downarrow$  & Ove$\downarrow$& FD$\downarrow$  & Rea$\downarrow$ & Occ$\downarrow$  \\
\midrule
\multirow{2}{*}{PKU}
&\multirow{2}{*}{CGL-hard}&PosterLlama
& \meanstd{0.0225}{0.0001} & \meanstd{0.0311}{0.0004}  & \meanstd{6.5679}{0.1091} & \meanstd{0.1758}{0.0004}  & \meanstd{0.0688}{0.0010} \\
&&\cellcolor{red!8}ReLayout (Ours) & \cellcolor{red!8}\textbf{\meanstd{0.0167}{0.0001}} & \cellcolor{red!8}\textbf{\meanstd{0.0100}{0.0004}}  & \cellcolor{red!8}\textbf{\meanstd{4.4413}{0.0136}} & \cellcolor{red!8}\textbf{\meanstd{0.1715}{0.0001}} & \cellcolor{red!8}\textbf{\meanstd{0.0631}{0.0002}}\\
\bottomrule
\multirow{2}{*}{CGL}
&\multirow{2}{*}{PKU-hard}&PosterLlama
& \meanstd{0.0019}{0.0007}  & \meanstd{0.0205}{0.0035} &  \meanstd{7.1093}{0.2796} & \textbf{\meanstd{0.1726}{0.0010}}  & \meanstd{0.0694}{0.0016} \\
&&\cellcolor{red!8}ReLayout (Ours) & \cellcolor{red!8}\textbf{\meanstd{0.0010}{0.0005}} & \cellcolor{red!8}\textbf{\meanstd{0.0120}{0.0023}} &  \cellcolor{red!8}\textbf{\meanstd{5.9011}{0.1120}} & \cellcolor{red!8}\meanstd{0.1730}{0.0012}  & \cellcolor{red!8}\textbf{\meanstd{0.0660}{0.0009}}\\
\bottomrule
    \end{tabular}
    }
    \label{tab:cross dataset}
\end{table*}

\begin{table*}[tb!]
\Large
    \centering
	\caption{Ablation study on the hard split of PKU dataset.}\label{tab:ablation}
 \resizebox{\textwidth}{!}{
	\begin{tabular}{lcccccccc}
		\toprule
		&{Region} &{Saliency} &{Resample} &$\Delta$Val$\downarrow$  & Ove$\downarrow$ &  FD$\downarrow$ & Rea$\downarrow$ & Occ$\downarrow$  \\
		\midrule
		V0&{-} & {-}  & {-} & \meanstd{0.0025}{0.0014} & \meanstd{0.0153}{0.0009} &  \meanstd{8.7960}{0.0224} & \textbf{\meanstd{0.1746}{0.0006}} & \meanstd{0.0821}{0.0006} \\
		V1&{\checkmark} & {-} & {-} & \meanstd{0.0021}{0.0002} & \meanstd{0.0379}{0.0013} & \meanstd{12.2290}{0.2255} & \meanstd{0.1967}{0.0005} & \meanstd{0.1188}{0.0011} \\
		V2&{\checkmark} & {\checkmark} & {-}& \meanstd{0.0014}{0.0007} & \meanstd{0.0150}{0.0019} & \meanstd{7.3406}{0.0719} & \meanstd{0.1769}{0.0002} & \meanstd{0.0754}{0.0014} \\
		V3&{\checkmark}& {\checkmark} & {\checkmark} & \cellcolor{red!8}{\textbf{\meanstd{0.0002}{0.0001}}} &  \cellcolor{red!8}\textbf{\meanstd{0.0097}{0.0004}} & \cellcolor{red!8}\textbf{\meanstd{4.9403}{0.1903}} & \cellcolor{red!8}\meanstd{0.1755}{0.0006} & \cellcolor{red!8}\textbf{\meanstd{0.0752}{0.0007}} \\
		\bottomrule
	\end{tabular}
 }
\end{table*}
\noindent\textbf{Qualitative comparison:} 
Figure~\ref{fig:big fig} visualizes the generated layouts, providing a comparison across different methods. 
It can be observed that, apart from the obvious errors marked in Figure~\ref{fig:big fig}, other methods also fall short in controlling element dimensions, maintaining spacing between elements, selecting layout arrangements, and achieving overall harmony. In contrast, ReLayout aligns more closely with human aesthetic preferences. Additionally, our method excels at generating diverse layouts and handling layout generation under various conditions, as demonstrated in the supplementary materials.

\noindent\textbf{Out-of-domain generalization:} 
To verify the generalization of our method, we conduct experiments using PKU as the training set and testing on CGL, and vice versa. As shown in Table~\ref{tab:cross dataset}, our method outperforms the current SOTA PosterLlama on most metrics. This demonstrates that ReLayout adapts well to real-world scenarios and demonstrates strong generalization performance.

\subsection{Ablation Study and Analysis}
\label{sec:ablation}
\noindent\textbf{Effect of Each Module:} We conduct a series of ablation experiments on the PKU test and hard split to evaluate the contributions of different modules in ReLayout. 
To simplify the setup, the ablation study uses a training set with only a single main condition from PKU: generating position and size given the category, text, and aspect ratio. 
As shown in Table~\ref{tab:ablation}, V1 adds only region annotations, V2 builds on V1 by incorporating salient annotations, and V3 further enhances the setup by introducing a layout prototype rebalance sampler, several observations can be made. 
First, V1 demonstrates relatively poor overall metrics, likely due to the model focusing on the structure of elements while neglecting salient objects. Since no structure-related metrics, this effect cannot be quantified and must be analyzed through visualization in the supplementary materials. Second, V2 shows that FD improves compared to direct fine-tuning (V0), especially in the hard split. Compared to the Region-only setting, all metrics show an upward trend, particularly the Occ metric, which demonstrates the importance of Saliency in content-aware tasks. Third, V3 shows the performance of ReLayout, demonstrating that it achieves the best results in the hard split across the Val, Ove, FD, and Occ metrics. Notably, the improvements in Ove and FD are particularly significant.
\newline
\noindent\textbf{Hyperparameter Analysis:} We analyze the hyperparameter $\theta$ on the hard split of PKU dataset, and the results are shown in Table~\ref{tab:hyper}. It can be observed that when $\theta = 6$, most of the metrics achieve their optimal values. This demonstrates that, when $\theta$ is smaller, the original distribution is largely preserved, which results in rare layout prototypes being treated as noise and thus receiving insufficient sampling. On the other hand, when $\theta$ is larger, the distribution shifts toward a more balanced state across groups, which causes certain rare layout prototypes to be repeatedly learned, thus degrading its performance.
\begin{table}[tb!]
\caption{Hyperparameter analysis.}
\Large
    \centering
    \resizebox{\columnwidth}{!}{
    \begin{tabular}{lccccc}
    \toprule
    $\theta$ & $\Delta$Val$\downarrow$  & Ove$\downarrow$ & FD$\downarrow$  & Rea$\downarrow$ & Occ$\downarrow$  \\
\midrule
3 & \meanstd{0.0009}{0.0004} & \meanstd{0.0121}{0.0001} &  \meanstd{4.7920}{0.1461} & \meanstd{0.1742}{0.0004} & \textbf{\meanstd{0.0633}{0.0006}} \\
\rowcolor{red!8}6& \textbf{\meanstd{0.0004}{0.0005}}& \textbf{\meanstd{0.0109}{0.0001}}& \textbf{\meanstd{3.4615}{0.1304}}& \textbf{\meanstd{0.1727}{0.0005}}& \meanstd{0.0637}{0.0002}\\
10 & \meanstd{0.0082}{0.0003} & \meanstd{0.0168}{0.0002} &  \meanstd{5.0655}{0.0347} & \meanstd{0.1791}{0.0005} & \meanstd{0.0703}{0.0006} \\
100 & \meanstd{0.0075}{0.0003} & \meanstd{0.0146}{0.0010} &  \meanstd{4.9287}{0.2307} & \meanstd{0.1783}{0.0001} & \meanstd{0.0808}{0.0011} \\
\bottomrule
    \end{tabular}
    }
    \label{tab:hyper}
\end{table}
\section{Conclusion}
In this work, we study content-aware layout generation tasks and address the issue in LLM-based methods where the relationships between elements have not been considered. We propose a novel method ReLayout, which consists of two modules. 
First, we enhance the model's understanding of relationships by incorporating explicit relationship annotations, framed from the perspective of CoT.
Second, we utilize relation annotations to cluster the dataset and adjust its distribution, thereby enhancing the quality of generated layouts.
Moreover, extensive experiments validate the effectiveness of our method, particularly in visualization.

Furthermore, we identify two limitations in ReLayout. 
Firstly, current metrics can identify obviously inadequate layouts, but they lack the ability to evaluate the suitability of layouts in real-world application scenarios. Furthermore, small changes in these metrics don't necessarily lead to a decline in layout quality.
Secondly, we have not applied relation annotations to other open-source MLLMs to verify whether this method is equally effective.

Future works can leverage our relation annotations for a more refined understanding in layout generation tasks. Additionally, we aim to maximize the effectiveness of our relation annotations through reinforcement learning in the future.
\newpage
\clearpage
\bibliography{aaai2026}

\begin{thebibliography}{43}
\providecommand{\natexlab}[1]{#1}

\bibitem[{Achiam et~al.(2023)Achiam, Adler, Agarwal, Ahmad, Akkaya, Aleman, Almeida, Altenschmidt, Altman, Anadkat et~al.}]{gpt}
Achiam, J.; Adler, S.; Agarwal, S.; Ahmad, L.; Akkaya, I.; Aleman, F.~L.; Almeida, D.; Altenschmidt, J.; Altman, S.; Anadkat, S.; et~al. 2023.
\newblock Gpt-4 technical report.
\newblock \emph{arXiv preprint arXiv:2303.08774}.

\bibitem[{Brown et~al.(2020)Brown, Mann, Ryder, Subbiah, Kaplan, Dhariwal, Neelakantan, Shyam, Sastry, Askell et~al.}]{fewshot}
Brown, T.; Mann, B.; Ryder, N.; Subbiah, M.; Kaplan, J.~D.; Dhariwal, P.; Neelakantan, A.; Shyam, P.; Sastry, G.; Askell, A.; et~al. 2020.
\newblock Language models are few-shot learners.
\newblock \emph{Advances in neural information processing systems}, 33: 1877--1901.

\bibitem[{Cai et~al.(2024)Cai, Cao, Chen, Chen, Chen, Chen, Chen, Chen, Chen, Chu et~al.}]{internlm}
Cai, Z.; Cao, M.; Chen, H.; Chen, K.; Chen, K.; Chen, X.; Chen, X.; Chen, Z.; Chen, Z.; Chu, P.; et~al. 2024.
\newblock Internlm2 technical report.
\newblock \emph{arXiv preprint arXiv:2403.17297}.

\bibitem[{Chai, Zhuang, and Yan(2023)}]{chai2023layoutdm}
Chai, S.; Zhuang, L.; and Yan, F. 2023.
\newblock Layoutdm: Transformer-based diffusion model for layout generation.
\newblock In \emph{Proceedings of the IEEE/CVF Conference on Computer Vision and Pattern Recognition}, 18349--18358.

\bibitem[{Chen et~al.(2024{\natexlab{a}})Chen, Wang, Cao, Liu, Gao, Cui, Zhu, Ye, Tian, Liu et~al.}]{internvl2.5}
Chen, Z.; Wang, W.; Cao, Y.; Liu, Y.; Gao, Z.; Cui, E.; Zhu, J.; Ye, S.; Tian, H.; Liu, Z.; et~al. 2024{\natexlab{a}}.
\newblock Expanding Performance Boundaries of Open-Source Multimodal Models with Model, Data, and Test-Time Scaling.
\newblock \emph{arXiv preprint arXiv:2412.05271}.

\bibitem[{Chen et~al.(2024{\natexlab{b}})Chen, Wu, Wang, Su, Chen, Xing, Zhong, Zhang, Zhu, Lu et~al.}]{internvit}
Chen, Z.; Wu, J.; Wang, W.; Su, W.; Chen, G.; Xing, S.; Zhong, M.; Zhang, Q.; Zhu, X.; Lu, L.; et~al. 2024{\natexlab{b}}.
\newblock Internvl: Scaling up vision foundation models and aligning for generic visual-linguistic tasks.
\newblock In \emph{Proceedings of the IEEE/CVF Conference on Computer Vision and Pattern Recognition}, 24185--24198.

\bibitem[{Deka et~al.(2017)Deka, Huang, Franzen, Hibschman, Afergan, Li, Nichols, and Kumar}]{UI2}
Deka, B.; Huang, Z.; Franzen, C.; Hibschman, J.; Afergan, D.; Li, Y.; Nichols, J.; and Kumar, R. 2017.
\newblock Rico: A mobile app dataset for building data-driven design applications.
\newblock In \emph{Proceedings of the 30th annual ACM symposium on user interface software and technology}, 845--854.

\bibitem[{Goodfellow et~al.(2020)Goodfellow, Pouget-Abadie, Mirza, Xu, Warde-Farley, Ozair, Courville, and Bengio}]{gan}
Goodfellow, I.; Pouget-Abadie, J.; Mirza, M.; Xu, B.; Warde-Farley, D.; Ozair, S.; Courville, A.; and Bengio, Y. 2020.
\newblock Generative adversarial networks.
\newblock \emph{Communications of the ACM}, 63(11): 139--144.

\bibitem[{Goyal et~al.(2024)Goyal, Rastogi, Rajagopal, Yuan, Zhao, Chintagunta, Naik, and Ward}]{healthcare}
Goyal, S.; Rastogi, E.; Rajagopal, S.~P.; Yuan, D.; Zhao, F.; Chintagunta, J.; Naik, G.; and Ward, J. 2024.
\newblock Healai: A healthcare llm for effective medical documentation.
\newblock In \emph{Proceedings of the 17th ACM International Conference on Web Search and Data Mining}, 1167--1168.

\bibitem[{Guo et~al.(2021)Guo, Jin, Sun, Li, Li, Shi, and Cao}]{poster1}
Guo, S.; Jin, Z.; Sun, F.; Li, J.; Li, Z.; Shi, Y.; and Cao, N. 2021.
\newblock Vinci: an intelligent graphic design system for generating advertising posters.
\newblock In \emph{Proceedings of the 2021 CHI conference on human factors in computing systems}, 1--17.

\bibitem[{Ho, Jain, and Abbeel(2020)}]{diffusion}
Ho, J.; Jain, A.; and Abbeel, P. 2020.
\newblock Denoising diffusion probabilistic models.
\newblock \emph{Advances in neural information processing systems}, 33: 6840--6851.

\bibitem[{Horita et~al.(2024)Horita, Inoue, Kikuchi, Yamaguchi, and Aizawa}]{ralf}
Horita, D.; Inoue, N.; Kikuchi, K.; Yamaguchi, K.; and Aizawa, K. 2024.
\newblock Retrieval-Augmented Layout Transformer for Content-Aware Layout Generation.
\newblock In \emph{Proceedings of the IEEE/CVF Conference on Computer Vision and Pattern Recognition}, 67--76.

\bibitem[{Hsu et~al.(2023)Hsu, He, Peng, Kong, and Zhang}]{pku}
Hsu, H.~Y.; He, X.; Peng, Y.; Kong, H.; and Zhang, Q. 2023.
\newblock Posterlayout: A new benchmark and approach for content-aware visual-textual presentation layout.
\newblock In \emph{Proceedings of the IEEE/CVF Conference on Computer Vision and Pattern Recognition}, 6018--6026.

\bibitem[{Hu et~al.(2022)Hu, Shen, Wallis, Allen-Zhu, Li, Wang, Wang, Chen et~al.}]{lora}
Hu, E.~J.; Shen, Y.; Wallis, P.; Allen-Zhu, Z.; Li, Y.; Wang, S.; Wang, L.; Chen, W.; et~al. 2022.
\newblock Lora: Low-rank adaptation of large language models.
\newblock \emph{ICLR}, 1(2): 3.

\bibitem[{Inoue et~al.(2023)Inoue, Kikuchi, Simo-Serra, Otani, and Yamaguchi}]{inoue2023layoutdm}
Inoue, N.; Kikuchi, K.; Simo-Serra, E.; Otani, M.; and Yamaguchi, K. 2023.
\newblock Layoutdm: Discrete diffusion model for controllable layout generation.
\newblock In \emph{Proceedings of the IEEE/CVF Conference on Computer Vision and Pattern Recognition}, 10167--10176.

\bibitem[{Jiang et~al.(2022)Jiang, Sun, Zhu, Lou, and Zhang}]{jiang2022coarse}
Jiang, Z.; Sun, S.; Zhu, J.; Lou, J.-G.; and Zhang, D. 2022.
\newblock Coarse-to-fine generative modeling for graphic layouts.
\newblock In \emph{Proceedings of the AAAI conference on artificial intelligence}, volume~36, 1096--1103.

\bibitem[{Jyothi et~al.(2019)Jyothi, Durand, He, Sigal, and Mori}]{jyothi2019layoutvae}
Jyothi, A.~A.; Durand, T.; He, J.; Sigal, L.; and Mori, G. 2019.
\newblock Layoutvae: Stochastic scene layout generation from a label set.
\newblock In \emph{Proceedings of the IEEE/CVF International Conference on Computer Vision}, 9895--9904.

\bibitem[{Kikuchi et~al.(2021)Kikuchi, Simo-Serra, Otani, and Yamaguchi}]{fid}
Kikuchi, K.; Simo-Serra, E.; Otani, M.; and Yamaguchi, K. 2021.
\newblock Constrained graphic layout generation via latent optimization.
\newblock In \emph{Proceedings of the 29th ACM International Conference on Multimedia}, 88--96.

\bibitem[{Kingma(2013)}]{vae}
Kingma, D.~P. 2013.
\newblock Auto-encoding variational bayes.
\newblock \emph{arXiv preprint arXiv:1312.6114}.

\bibitem[{Li et~al.(2024)Li, Zhang, Guo, Zhang, Li, Zhang, Zhang, Zhang, Li, Liu et~al.}]{llavaov}
Li, B.; Zhang, Y.; Guo, D.; Zhang, R.; Li, F.; Zhang, H.; Zhang, K.; Zhang, P.; Li, Y.; Liu, Z.; et~al. 2024.
\newblock Llava-onevision: Easy visual task transfer.
\newblock \emph{arXiv preprint arXiv:2408.03326}.

\bibitem[{Li et~al.(2023{\natexlab{a}})Li, Liu, Feng, Zhu, Li, Zhang, Lv, Zhu, Shen, Lin et~al.}]{radm}
Li, F.; Liu, A.; Feng, W.; Zhu, H.; Li, Y.; Zhang, Z.; Lv, J.; Zhu, X.; Shen, J.; Lin, Z.; et~al. 2023{\natexlab{a}}.
\newblock Relation-aware diffusion model for controllable poster layout generation.
\newblock In \emph{Proceedings of the 32nd ACM International Conference on Information and Knowledge Management}, 1249--1258.

\bibitem[{Li et~al.(2023{\natexlab{b}})Li, Li, Savarese, and Hoi}]{blip}
Li, J.; Li, D.; Savarese, S.; and Hoi, S. 2023{\natexlab{b}}.
\newblock Blip-2: Bootstrapping language-image pre-training with frozen image encoders and large language models.
\newblock In \emph{International conference on machine learning}, 19730--19742. PMLR.

\bibitem[{Li et~al.(2019{\natexlab{a}})Li, Yang, Hertzmann, Zhang, and Xu}]{document1}
Li, J.; Yang, J.; Hertzmann, A.; Zhang, J.; and Xu, T. 2019{\natexlab{a}}.
\newblock LayoutGAN: Generating Graphic Layouts with Wireframe Discriminators.
\newblock In \emph{International Conference on Learning Representations}.

\bibitem[{Li et~al.(2019{\natexlab{b}})Li, Yang, Hertzmann, Zhang, and Xu}]{li2019layoutgan}
Li, J.; Yang, J.; Hertzmann, A.; Zhang, J.; and Xu, T. 2019{\natexlab{b}}.
\newblock Layoutgan: Generating graphic layouts with wireframe discriminators.
\newblock \emph{arXiv preprint arXiv:1901.06767}.

\bibitem[{Lin et~al.(2024)Lin, Guo, Sun, Yang, Lou, and Zhang}]{layoutprompter}
Lin, J.; Guo, J.; Sun, S.; Yang, Z.; Lou, J.-G.; and Zhang, D. 2024.
\newblock Layoutprompter: Awaken the design ability of large language models.
\newblock \emph{Advances in Neural Information Processing Systems}, 36.

\bibitem[{Lin et~al.(2023)Lin, Zhou, Ma, Gao, Fei, Chen, Yu, and Ge}]{poster2}
Lin, J.; Zhou, M.; Ma, Y.; Gao, Y.; Fei, C.; Chen, Y.; Yu, Z.; and Ge, T. 2023.
\newblock Autoposter: A highly automatic and content-aware design system for advertising poster generation.
\newblock In \emph{Proceedings of the 31st ACM International Conference on Multimedia}, 1250--1260.

\bibitem[{Peng et~al.(2023)Peng, Liu, Yang, Yuan, and Li}]{agriculture}
Peng, R.; Liu, K.; Yang, P.; Yuan, Z.; and Li, S. 2023.
\newblock Embedding-based retrieval with llm for effective agriculture information extracting from unstructured data.
\newblock \emph{arXiv preprint arXiv:2308.03107}.

\bibitem[{Radford et~al.(2021)Radford, Kim, Hallacy, Ramesh, Goh, Agarwal, Sastry, Askell, Mishkin, Clark et~al.}]{clip}
Radford, A.; Kim, J.~W.; Hallacy, C.; Ramesh, A.; Goh, G.; Agarwal, S.; Sastry, G.; Askell, A.; Mishkin, P.; Clark, J.; et~al. 2021.
\newblock Learning transferable visual models from natural language supervision.
\newblock In \emph{International conference on machine learning}, 8748--8763. PmLR.

\bibitem[{Raneburger, Popp, and Vanderdonckt(2012)}]{UI1}
Raneburger, D.; Popp, R.; and Vanderdonckt, J. 2012.
\newblock An automated layout approach for model-driven WIMP-UI generation.
\newblock In \emph{Proceedings of the 4th ACM SIGCHI symposium on Engineering interactive computing systems}, 91--100.

\bibitem[{Seol, Kim, and Yoo(2024)}]{posterllama}
Seol, J.; Kim, S.; and Yoo, J. 2024.
\newblock PosterLlama: Bridging Design Ability of Language Model to Content-Aware Layout Generation.
\newblock In \emph{European Conference on Computer Vision}, 451--468. Springer.

\bibitem[{Tabata et~al.(2019)Tabata, Yoshihara, Maeda, and Yokoyama}]{magazines2}
Tabata, S.; Yoshihara, H.; Maeda, H.; and Yokoyama, K. 2019.
\newblock Automatic layout generation for graphical design magazines.
\newblock In \emph{ACM SIGGRAPH 2019 Posters}, 1--2.

\bibitem[{Tang et~al.()Tang, Wu, Li, and Duan}]{layoutnuwa}
Tang, Z.; Wu, C.; Li, J.; and Duan, N. ????
\newblock LayoutNUWA: Revealing the Hidden Layout Expertise of Large Language Models.
\newblock In \emph{The Twelfth International Conference on Learning Representations}.

\bibitem[{Team et~al.(2023)Team, Anil, Borgeaud, Alayrac, Yu, Soricut, Schalkwyk, Dai, Hauth, Millican et~al.}]{gemini}
Team, G.; Anil, R.; Borgeaud, S.; Alayrac, J.-B.; Yu, J.; Soricut, R.; Schalkwyk, J.; Dai, A.~M.; Hauth, A.; Millican, K.; et~al. 2023.
\newblock Gemini: a family of highly capable multimodal models.
\newblock \emph{arXiv preprint arXiv:2312.11805}.

\bibitem[{Tzachor et~al.(2023)Tzachor, Devare, Richards, Pypers, Ghosh, Koo, Johal, and King}]{agriculture1}
Tzachor, A.; Devare, M.; Richards, C.; Pypers, P.; Ghosh, A.; Koo, J.; Johal, S.; and King, B. 2023.
\newblock Large language models and agricultural extension services.
\newblock \emph{Nature food}, 4(11): 941--948.

\bibitem[{Vaswani(2017)}]{transformer}
Vaswani, A. 2017.
\newblock Attention is all you need.
\newblock \emph{Advances in Neural Information Processing Systems}.

\bibitem[{Wei et~al.(2022)Wei, Wang, Schuurmans, Bosma, Xia, Chi, Le, Zhou et~al.}]{cot}
Wei, J.; Wang, X.; Schuurmans, D.; Bosma, M.; Xia, F.; Chi, E.; Le, Q.~V.; Zhou, D.; et~al. 2022.
\newblock Chain-of-thought prompting elicits reasoning in large language models.
\newblock \emph{Advances in neural information processing systems}, 35: 24824--24837.

\bibitem[{Yang et~al.(2016)Yang, Mei, Xu, Rui, and Li}]{magazines1}
Yang, X.; Mei, T.; Xu, Y.-Q.; Rui, Y.; and Li, S. 2016.
\newblock Automatic generation of visual-textual presentation layout.
\newblock \emph{ACM Transactions on Multimedia Computing, Communications, and Applications (TOMM)}, 12(2): 1--22.

\bibitem[{Yang et~al.(2024)Yang, Xu, Yao, Rogers, Zhang, Intille, Shara, Gao, and Wang}]{healthcare1}
Yang, Z.; Xu, X.; Yao, B.; Rogers, E.; Zhang, S.; Intille, S.; Shara, N.; Gao, G.~G.; and Wang, D. 2024.
\newblock Talk2care: An llm-based voice assistant for communication between healthcare providers and older adults.
\newblock \emph{Proceedings of the ACM on Interactive, Mobile, Wearable and Ubiquitous Technologies}, 8(2): 1--35.

\bibitem[{Yu et~al.(2020)Yu, Gong, Jiang, Ye, and Han}]{yu2020scale}
Yu, X.; Gong, Y.; Jiang, N.; Ye, Q.; and Han, Z. 2020.
\newblock Scale match for tiny person detection.
\newblock In \emph{Proceedings of the IEEE/CVF winter conference on applications of computer vision}, 1257--1265.

\bibitem[{Zhang et~al.(2023)Zhang, Guo, Sun, Lou, and Zhang}]{zhang2023layoutdiffusion}
Zhang, J.; Guo, J.; Sun, S.; Lou, J.-G.; and Zhang, D. 2023.
\newblock Layoutdiffusion: Improving graphic layout generation by discrete diffusion probabilistic models.
\newblock In \emph{Proceedings of the IEEE/CVF International Conference on Computer Vision}, 7226--7236.

\bibitem[{Zheng et~al.(2019)Zheng, Qiao, Cao, and Lau}]{contentgan}
Zheng, X.; Qiao, X.; Cao, Y.; and Lau, R.~W. 2019.
\newblock Content-aware generative modeling of graphic design layouts.
\newblock \emph{ACM Transactions on Graphics (TOG)}, 38(4): 1--15.

\bibitem[{Zhong, Tang, and Yepes(2019)}]{document2}
Zhong, X.; Tang, J.; and Yepes, A.~J. 2019.
\newblock Publaynet: largest dataset ever for document layout analysis.
\newblock In \emph{2019 International conference on document analysis and recognition (ICDAR)}, 1015--1022. IEEE.

\bibitem[{Zhou et~al.(2022)Zhou, Xu, Ma, Ge, Jiang, and Xu}]{cgl}
Zhou, M.; Xu, C.; Ma, Y.; Ge, T.; Jiang, Y.; and Xu, W. 2022.
\newblock Composition-aware graphic layout GAN for visual-textual presentation designs.
\newblock In \emph{IJCAI}.

\end{thebibliography}

\end{document}